\documentclass[hf]{ceurart}

\makeatletter
\expandafter\def\csname msg_term:n\endcsname#1{}
\makeatother

\makeatletter
\ExplSyntaxOn
\RenewDocumentCommand \fntext { O{} m }
{
  \tl_if_head_eq_catcode:nNTF { #1 } a
  {
    \seq_gput_right:Nn \g_ceur_fnote_seq
    { \int_incr:N \g_ceur_fnote_int
      \str_set:Nx \@currentlabel { \int_use:N \g_ceur_fnote_int }
      \ceurLabel { #1 }
      \tex_def:D \thefootnote
      { \int_case:nn { \g_ceur_fnote_int } { { 1 } { $\dagger$ } { 2 } { $\ddagger$ } } }
      \footnotetext { #2 }
    }
  }
  {
    \seq_gput_right:Nn \g_ceur_fnote_seq
    { \int_set:Nn \l_tmpa_int { #1 }
      \tex_def:D \thefootnote
      { \int_case:nn { \l_tmpa_int } { { 1 } { $\dagger$ } { 2 } { $\ddagger$ } } }
      \footnotetext { #2 }
    }
  }
}
\RenewDocumentCommand \process@marks { }
{
  \cs_if_free:cTF { mark@corau\theauthor }
    { \ignorespaces }
    { \str_set:Nx \l_tmpa_str { \use:c{ mark@corau\theauthor } }
      \int_case:nn { \l_tmpa_str }
        { { 1 } { \sep$\ast$ } { 2 } { \sep$\ast\ast$ } { 3 } { \sep$\ast\!\ast\!\ast$ } }
      \tex_def:D \sep{\unskip,}
    }
  \cs_if_free:cTF { mark@fnau\theauthor }
    { \ignorespaces }
    { \str_set:Nx \l_tmpa_str { \use:c{ mark@fnau\theauthor } }
      \int_case:nn { \l_tmpa_str }
        { { 1 } { \sep$\dagger$ } { 2 } { \sep$\ddagger$ } }
      \tex_def:D \sep{\unskip,}
    }
}
\ExplSyntaxOff
\makeatother

\usepackage{minted}
\usepackage{makecell}
\setminted{breaklines=true}

\usepackage{graphicx}

\usepackage{hyperref}
\usepackage{array}
\usepackage{multirow}
\usepackage{color, colortbl}
\usepackage{subcaption}
\usepackage{arydshln}

\usepackage[table]{xcolor}   
\usepackage{tabularx}

\usepackage{xspace}

\newcommand{\rparagraph}[1]{\vspace{1.2mm}\noindent\textbf{#1.}}

\usepackage{todonotes}

\definecolor{Gray}{gray}{0.9}
\definecolor{ashgrey}{rgb}{0.7, 0.75, 0.71}
\definecolor{darkgreen}{rgb}{0,0.39,0}

\newcolumntype{g}{>{\columncolor{Gray}}c}

\newcommand{\kg}{KG\xspace}
\newcommand{\kgs}{KGs\xspace}
\newcommand{\llm}{LLM\xspace}
\newcommand{\llms}{LLMs\xspace}
\newcommand{\ctd}{CTD\xspace}

\newcommand{\lm}{LM\xspace}
\newcommand{\lms}{LMs\xspace}
\newcommand{\pubmed}{PubMed\xspace}
\newcommand{\ctdalign}{CTD-Align\xspace}

\begin{document}

\copyrightyear{2026}
\copyrightclause{Copyright for this paper by its authors.
  Use permitted under Creative Commons License Attribution 4.0
  International (CC BY 4.0).}

\conference{Third Workshop on Knowledge Graphs and Neurosymbolic AI (KG-NeSy 2026), in ISWC 2026 Workshops Joint Proceedings, October 25--26, 2026, Bari, Italy}

\title{Aligning Biomedical Texts and Knowledge Graphs: A Systematic Comparison of Lightweight Alignment Strategies}

\author[1]{Artem Bisliouk}[%
email=artem.bislyuk@students.uni-mannheim.de
]
\cormark[1]

\author[1]{Elizaveta Nosova}[%
email=elizaveta.nosova@students.uni-mannheim.de
]
\cormark[1]

\author[2]{Heiko Paulheim}[%
orcid=0000-0003-4386-8195,
email=heiko.paulheim@uni-mannheim.de
]

\author[2]{Andreea Iana}[%
orcid=0000-0002-7248-7503,
email=andreea.iana@uni-mannheim.de
]
\cormark[1] 
\fnmark[1]

\author[2]{Rita T. Sousa}[%
orcid=0000-0002-7241-8970,
email=rita.sousa@uni-mannheim.de
] \cormark[1]

\address[1]{University of Mannheim, Germany}
\address[2]{Data and Web Science Group, University of Mannheim, Germany}

\cortext[1]{Equal contribution.}
\fntext[1]{Corresponding author.}

\begin{abstract}
  Biomedical knowledge exists in two complementary but distinct forms: unstructured scientific literature and structured knowledge graphs (\kgs). Aligning them is essential for knowledge grounding, evidence retrieval, and \kg completion, yet existing methods do not explicitly align free-text evidence with \kg triples. 
  We present a unified framework for systematically studying design choices for aligning biomedical text and \kgs. With a text encoder and a \kg embedding model both frozen, we learn only a lightweight projection between their spaces via a contrastive objective. This enables a fair comparison across six design dimensions: text encoder, \kg embedding model, projection head, triple composition, training direction, and hard-negatives sampling.
  We construct \textit{\ctdalign}, a corpus of over 22K one-to-one triple-document pairs linking chemical-gene interactions from the Comparative Toxicogenomics Database to supporting PubMed passages.
  We evaluate alignment on it in two retrieval settings: document-to-triple and triple-to-document. We find that the triple composition and the training direction (i.e., shared retrieval space) have the greatest impact, whereas the text encoder and hard-negatives sampling matter little. Overall, simple choices win: projecting text into the \kg space with a linear head over concatenated subject, predicate, and object embeddings performs best. 
  These findings establish lightweight contrastive alignment as an effective, practical foundation for bridging biomedical text and \kgs.
\end{abstract}


\maketitle

\section{Introduction}
\label{sec:introduction}

Large language models (\llms) have showcased impressive natural language understanding and generation capabilities \cite{brown2020language,wei2022emergent,lievin2024can,zhao2023survey}, making them increasingly valuable for advancing domains characterized by massive amounts of complex, unstructured data \cite{luo2022biogpt,thirunavukarasu2023large,zhou2025large}. Despite their fluency, however, \llms remain prone to hallucinating convincing yet unsupported or factually incorrect answers \cite{ji2023survey,huang2025survey} and to relying on outdated knowledge \cite{kasai2023realtime,zhu2025your}. These limitations are particularly concerning in knowledge-intensive domains that require specialized expertise \cite{mallen2023not}, such as biomedicine and healthcare, where incorrect claims about chemicals, genes, or diseases may have direct clinical and toxicological consequences \cite{omar2025multi,singhal2023large,tian2024opportunities}.  

Biomedical knowledge is available in two complementary forms. 
The biomedical literature, through repositories such as \pubmed \cite{lu2011pubmed}, PMC \cite{roberts2001pubmed}, or Europe PMC \cite{europe2015europe}, provides a vast but unstructured collection of scientific publications that \llms can easily process. 
In parallel, biomedical knowledge graphs (\kgs) provide structured, manually curated, and continually updated facts about biomedical entities (e.g., chemicals, genes, diseases) and their relationships \cite{himmelstein2017systematic,nicholson2020constructing}. 
Although these resources describe many overlapping facts, they represent them fundamentally differently: biomedical literature expresses knowledge in natural language, whereas \kgs encode it as symbolic triples. Consequently, there is no explicit correspondence between a scientific document (or one of its sections) and the relation that it describes. Yet many biomedical applications, such as grounding \llm-generated responses in \kg information to mitigate hallucinations \cite{pan2024unifying,soman2024biomedical,matsumoto2024kragen}, retrieving textual evidence supporting or contradicting a given \kg fact \cite[\emph{inter alia}]{syed2018factcheck,zhang2022distant,wuhrl2024makes}, and extracting relations from the literature to enrich or complete \kgs as new findings are published \cite{mintz2009distant,miranda2023overview,wei2024pubtator}, require precisely such correspondences.
However, existing methods do not explicitly align biomedical free-text evidence with \kg triples.

Existing approaches for integrating language models (\lms) and \kgs suffer from two major limitations.
First, representation integration approaches either inject pre-aligned \kg entity or triple embeddings into an \lm to enrich its representations \cite{zhang2019ernie,peters2019knowledge,liu2020k,wang2021kepler} or leverage \lms for \kg-specific tasks, such as link prediction or \kg completion, by encoding entities and relations using their textual descriptions \cite{toutanova2015representing,yao2019kgbert,wang2022simkgc}. Consequently, these methods are optimized for downstream tasks (i.e., question answering, relation extraction), rather than for learning an explicit mapping between free-text evidence and structured triples. 
Second, text-graph alignment methods typically align individual nodes instead of complete facts \cite{kartsaklis2018mapping,brannon2024congrat} or ignore graph structure by linearizing triples into text \cite{le2023joint}. Moreover, these techniques are developed and evaluated on general-domain data, whose vocabulary, entity types, and relation semantics differ significantly from those found in biomedicine \cite{gu2021domain, gururangan2020don}. 

In this work, we address these shortcomings by investigating which design choices drive the alignment quality between structured and unstructured representations of biomedical facts. 
Concretely, we make three main contributions.
\textbf{1)} We introduce a unified, modular framework for systematically comparing alignment strategies between \lms and \kgs. The framework is built around a simple backbone: given a biomedical text encoder and a \kg embedding model, we keep both components \textit{frozen} and train only a lightweight projection between their embedding spaces using a contrastive (InfoNCE) objective \cite{oord2018representation}. This training objective encourages paired documents and triples to occupy nearby positions in a shared embedding space. 
We evaluate alignment quality in two retrieval settings: recovering the \kg triple expressed by a document (\textit{document-to-triple}) and, conversely, retrieving the document describing a triple (\textit{triple-to-document}).
Keeping the backbone fixed, we isolate the impact of six key design choices: the text encoder, the \kg embedding model, the projection head, the triple-composition operator, the training direction, and the hard-negatives sampling strategy. By varying one component at a time under a common tuning and evaluation protocol we can attribute performance differences directly to each design dimension. 
\textbf{2)} To the best of our knowledge, no existing resource pairs biomedical triples with the documents that express them. We therefore construct \textit{\ctdalign}, a reusable corpus of 22,293 one-to-one triple-document pairs with full provenance, obtained by pairing chemical-gene interactions from the Comparative Toxicogenomics Database (CTD) \cite{davis2025comparative} with their supporting evidence from \pubmed articles \cite{lu2011pubmed}. The resulting dataset supports future research on text-graph alignment, evidence retrieval, and knowledge attribution.
\textbf{3)} Lastly, we find that the triple-composition operator and the training direction (i.e., the shared retrieval space) have the greatest impact on alignment quality, whereas the choice of text encoder and hard-negatives sampling strategy has little impact once the spaces are aligned. We further show that simpler design choices outperform more complex alternatives. In particular, projecting text into the \kg space with a linear head over concatenated subject, predicate, and object embeddings yields the best alignment in both retrieval settings. 

Overall, our findings demonstrate that lightweight alignment constitutes an effective bridge between biomedical text and \kg representations, offering a simple yet practical foundation for applications such as knowledge grounding, evidence retrieval, and \kg completion.
\section{Related Work}
\label{sec:related_work}

\subsection{Integrating Language Models and Knowledge Graphs}
\label{sec:lm_kg_integration}

Integrating \lms and \kgs constitutes a well-established research direction \cite{hu2023survey,pan2024unifying}.
One line of work injects structured knowledge into \lms to improve downstream tasks such as relation classification or question answering: ERNIE \cite{zhang2019ernie} and KnowBERT \cite{peters2019knowledge} fuse linked entity embeddings into contextual word representations, K-BERT \cite{liu2020k} inserts relevant triples into the input as a soft-positioned sentence tree, KEPLER \cite{wang2021kepler} jointly optimizes a knowledge embedding and a masked language modeling objective, and DRAGON \cite{yasunaga2022deep} fuses a text encoder and a graph neural network through joint pretraining.
A complementary line uses \lms for \kg-specific tasks, most prominently \kg completion. Early works learned a shared text-\kg embedding space from scratch, aligning word and entity embeddings through co-occurrence and entity names \cite{wang2014knowledge,toutanova2015representing} or encoding entity descriptions directly into the \kg model \cite{xie2016representation}. Later approaches, such as KG-BERT \cite{yao2019kgbert}, fine-tune BERT as a cross-encoder over a triple's concatenated component descriptions to score its plausibility, whereas SimKGC \cite{wang2022simkgc} recasts this as a contrastive learning with a bi-encoder for efficient and more accurate \kg completion.

All of these approaches primarily train the underlying encoders for a downstream task or for \kg completion. We instead keep both the \kg and the text encoder \textit{frozen}, learn only a lightweight projection between their output spaces, and target the retrieval between evidence documents and triples rather than knowledge injection or completion. 

\subsection{Embedding Space Alignment and Retrieval}
\label{sec:emb_space_alignment}

We study an alignment scheme from a broader family that maps two independently trained embedding spaces onto one another with a lightweight transformation, an idea rooted in cross-lingual work, where a linear map aligns separately trained word embeddings \cite{mikolov2013exploiting}. 
The same principle underpins cross-modal alignment. CLIP \cite{radford2021learning} aligns image and text encoders with an InfoNCE objective \cite{oord2018representation}, LiT \cite{zhai2022lit} shows that freezing one encoder and tuning only the projection often suffices, and recent work aligns two frozen unimodal encoders by training only lightweight projectors, matching contrastively trained models at a fraction of the data and compute \cite{maniparambil2025harnessing}.

Several works align natural language with graph structure. 
ConGraT \cite{brannon2024congrat} contrastively aligns node-associated text with graph representations, and JointGT \cite{ke2021jointgt} learns joint graph-text representations with an alignment objective for \kg-to-text generation. In contrast to our approach, these models train their encoders rather than keeping them frozen, and align individual nodes or general graph-structured inputs rather than \kg triples.
A similar approach is proposed by \citep{le2023joint}, who linearize RDF graphs into token sequences and train twin Transformer encoders with an in-batch contrastive loss so that a document and its parallel graph become neighbors.
Our setting differs in three ways: we retain an independently trained structural \kg embedding space instead of re-encoding a linearized graph as text and learn only a lightweight projection over frozen encoders. Moreover, we pair each document with an exact subject-predicate-object triple rather than with a graph that may contain multiple triples.

\subsection{Text-Knowledge Graph Alignment in the Biomedical Domain}
\label{sec:biomedical_alignment}

A key motivation for aligning text and \kgs is the tendency of \llms to hallucinate \cite{ji2023survey}. Retrieval-augmented generation grounds \llm outputs in retrieved documents \cite{lewis2020retrieval}. Biomedical variants extend this idea to structured sources such as \kgs for multi-hop reasoning and prompt construction \cite{matsumoto2024kragen,soman2024biomedical}, drawing on curated resources such as \ctd \cite{davis2025comparative}, Hetionet \cite{himmelstein2017systematic}, and PrimeKG \cite{chandak2023building}.
Aligned-LLM \cite{nishat2025aligning} aligns frozen \kg entity embeddings with an \lm's space via a learned projection to improve factual grounding, and BALI \cite{sakhovskiy2025bali} contrastively aligns an \lm with a UMLS graph encoder using linked concept mentions. At the mention level, biomedical bi-encoders such as SapBERT \cite{liu2021self} link text spans to single \kg concepts. 
These approaches operate at the entity or concept level and target improved generation or classification. In contrast, we align triples with their evidence documents, evaluate bidirectional retrieval between them, and study the effect of different encoders and alignment components under a controlled frozen-encoder framework. To the best of our knowledge, such a systematic study of triple-level biomedical text-\kg alignment has not yet been conducted. 

\section{Methodology}
\label{sec:methodology}

\begin{figure}
    \centering
    \includegraphics[width=\linewidth]{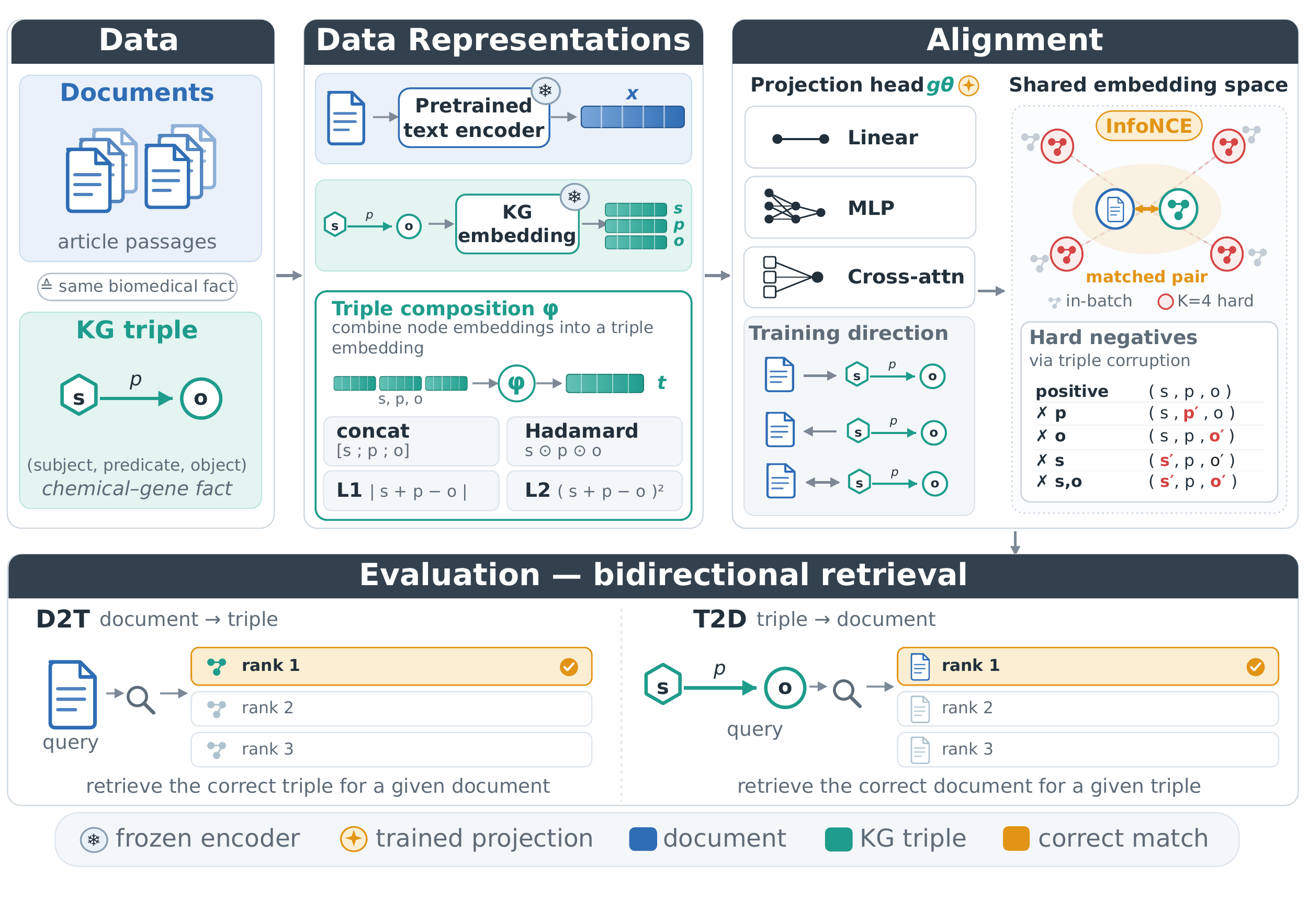}
    \caption{Overview of our unified framework for biomedical text-\kg alignment. Documents and \kg triples expressing the same fact are embedded by frozen encoders; the only trained component is a lightweight projection aligning the two spaces under a contrastive objective. We vary six components, shown here as separate modules: the text encoder, \kg embedding model, triple-composition operator, projection head, training direction, and hard-negative strategy. We evaluate alignment by retrieval in two settings: document-to-triple (\texttt{D2T}) and triple-to-document (\texttt{T2D}). 
    }
    \label{fig:methodology}
\end{figure}

Fig.~\ref{fig:methodology} depicts our unified framework for evaluating biomedical text-\kg alignment along six critical dimensions that determine its quality. Given biomedical documents and \kg triples describing the same facts, we first encode each modality with a dedicated text encoder and \kg embedding model (\S\ref{sec:text_kg_embeddings}). Keeping both encoders frozen, we then learn only a lightweight projection between their embedding spaces using a contrastive objective, varying the training direction and hard-negatives sampling strategy  (\S\ref{sec:alignment}). Because both encoders remain frozen, the framework is modular: any pretrained text or \kg encoder can be substituted, and only the alignment must be retrained, at a fraction of the cost of fine-tuning either encoder.
Finally, we assess alignment quality through bidirectional retrieval between documents and triples. 
Next, we describe each dimension and formalize the concrete design choices.

\subsection{Data Representations}
\label{sec:text_kg_embeddings}

The input data consists of biomedical facts, each expressed in two complementary modalities.
In a knowledge graph $G \subseteq E \times R \times E$ with entity set $E$ and relation set $R$, a fact is represented as a triple $t=(s, p, o)$ with subject and object entities $s, o \in E$ and a predicate (relation) $p \in R$. 
In the biomedical literature, the same fact is described by a document $x_i$, namely a passage of an article (i.e., sentence or paragraph), that states the relation between $s$ and $o$. 
The input data thus consists of aligned pairs $\mathcal{P} = \{(x_i, t_i)\}_{i=1}^n$, where document $x_i$ expresses triple $t_i$.
We embed the two modalities separately with a \textit{frozen} text encoder and a \textit{frozen} \kg embedding model: each document is mapped to $\mathbf{x}_i \in \mathbb{R}^{d_\text{txt}}$, and its triple to $\mathbf{t}_i = \phi(\mathbf{s}_i, \mathbf{p}_i, \mathbf{o}_i) \in \mathbb{R}^{d_\text{kg}}$, where  $\mathbf{s}_i, \mathbf{p}_i, \mathbf{o}_i$ are the \kg embeddings of $s_i, p_i, o_i$ of dimensionality $d$, and $\phi$ combines them into a single representation.
As the choice of $\phi$ constitutes an important design dimension, we evaluate four triple-composition operators: (i) \textit{concatenation} $\mathbf{t}_i = [\mathbf{s}_i; \mathbf{p}_i; \mathbf{o}_i]$, (ii) \textit{Hadamard product} $\mathbf{t}_i = \mathbf{s}_i \odot \mathbf{p}_i \odot \mathbf{o}_i$, (iii) element-wise \textit{L\textsubscript{1}} combination $\mathbf{t}_i = \left| \mathbf{s}_i + \mathbf{p}_i - \mathbf{o}_i \right|$, and (iv) element-wise \textit{L\textsubscript{2}} combination $\mathbf{t}_i = (\mathbf{s}_i + \mathbf{p}_i - \mathbf{o}_i)^2$.

\subsection{Alignment Methods}
\label{sec:alignment}

Starting from the frozen document and triple embeddings obtained in the previous step, the alignment stage learns a lightweight projection $g_\theta$ that maps one modality into the embedding space of the other, such that a document and its corresponding triple become nearest neighbors under cosine similarity.
The triple-composition operator $\phi$ determines the triple embedding dimension: \textit{concatenation} yields $d_\text{kg} = 3d$, whereas the \textit{Hadamard product} and the \textit{L\textsubscript{1}} and \textit{L\textsubscript{2}} approaches yield $d_\text{kg} = d$. The projection therefore maps between $\mathbb{R}^{d_\text{txt}}$ and $\mathbb{R}^{d_\text{kg}}$, with the input and output sizes set by the training direction and the triple-composition operator.

\rparagraph{Projection Heads}
We compare three projection heads of increasing complexity. 
The \textit{Linear} head applies a single linear transformation without any hidden layer, nonlinearity, normalization, and dropout. 
The \textit{MLP} head is a feed-forward network whose hidden layers each apply a linear transformation, layer normalization, a GELU activation, and dropout, before 
a final linear layer maps the intermediate representation to the target dimension.
Finally, the \textit{cross-attention} head projects the document embedding into a single key-value vector, to which three learnable query vectors -- one per triple element (subject, predicate, object) -- attend through a multi-head attention layer. Their outputs are concatenated and linearly mapped to the target dimension. As the key-value sequence is a single vector, this amounts to a learned multi-query projection of the document embedding rather than token-level attention over words to triple components. 

\rparagraph{Training Direction}
The training direction determines which embedding space serves as the alignment target. We consider three settings: (i) \textit{text$\to$\kg} projects document embeddings into the \kg triple space, (ii) \textit{\kg$\to$text} projects triple embeddings into the document space, and (iii) \textit{bidirectional} randomly assigns each batch to one of the two directions. As the input and output dimensionalities differ, we train a separate projection network for each direction. 

\rparagraph{Training Objective}
We train all alignment models with the InfoNCE loss \cite{oord2018representation}. For each anchor (a document or triple), its pair is the positive and the other mini-batch items are in-batch negatives. 
Optionally, we augment these with hard negatives by corrupting a positive triple $(s, p, o)$ in the \kg space.
We compare four corruption schemes that replace (i) the \textit{predicate} (keeping $s$ and $o$), (ii) the \textit{object} (keeping $s$ and $p$), (iii) the \textit{subject} (keeping $p$ and $o$), or (iv) \textit{both entities} (keeping only the predicate). We also consider using no hard negatives (\textit{None}, i.e., in-batch negatives only) and the union of all four schemes (\textit{All}).
Because the corruptions live in the \kg-triple space, they apply naturally to the text$\to$\kg training direction, where the model predicts a triple embedding that the loss contrasts against the true triple and its corruptions. In the \textit{\kg$\to$text} direction the target is a document embedding, for which no hard negatives are generated. In the bidirectional setting we use hard negatives only on text$\to$\kg batches.

\subsection{Retrieval Tasks}
\label{sec:}

We evaluate alignment quality as retrieval over the full candidate pool. Given a query (i.e., a document $x_i$ or a triple $t_i$), we project it through the learned mapping and rank all candidates of the target modality by cosine similarity.
We thus define two complementary tasks: (i) \textit{document-to-triple} (\texttt{D2T}), retrieving the correct \kg triple for a given evidence document, and (ii) \textit{triple-to-document} (\texttt{T2D}), retrieving the correct document for a given triple.

\section{Biomedical Corpora \& Dataset Construction}
\label{sec:corpora}

Our analysis relies on biomedical facts expressed in two complementary forms: as structured \kg triples and as free-text evidence. We take the structured facts from the Comparative Toxicogenomics Database knowledge graph (\S\ref{sec:ctd_kg}) and the textual evidence from PubMed (\S\ref{sec:pubmed_corpus}), and combine them into a new triple-document dataset \textit{\ctdalign} (\S\ref{sec:ctdalign}).

\subsection{CTD Knowledge Graph}
\label{sec:ctd_kg}
The Comparative Toxicogenomics Database (CTD) is a publicly available resource that manually curates relationships between chemicals, genes, phenotypes, and human diseases from the scientific literature  \cite{mattingly2006comparative,davis2025comparative}. It is part of the Bio2RDF \cite{belleau2008bio2rdf} linked data project for the life sciences and comprises roughly 3.8 million curated direct interactions, from which further relationships are inferred.
Crucially for our purpose, every direct interaction is annotated with the PubMed identifier(s) of the article(s) that report it (PMIDs), giving each fact an explicit textual provenance.

We extract several association types over four entity classes (chemicals, genes, diseases, pathways): (i) chemical-gene interactions (restricted to single-action interactions), (ii) chemical-disease and gene-disease associations (both restricted to manually curated direct evidence and flattened to \textit{treats} and \textit{marker/mechanism} relations), and (iii) gene-pathway memberships (i.e., \textit{is participant in}).
Chemical-gene interactions form the backbone of the \textit{core} \kg and are the only interactions paired with documents, whereas the remaining associations enrich the graph's structure to provide the context needed to learn informative \kg embeddings.

\rparagraph{Dereification}
Like many other RDF resources, \ctd represents each n-ary interaction through reification, i.e., an intermediate node to which metadata can be attached.\footnote{A single chemical-gene interaction is encoded as three triples: (\textit{chemical}$\to$\textit{interaction\_node}), (\textit{interaction\_node}$\to$\textit{action}), and (\textit{interaction\_node}$\to$\textit{gene}).}
Standard \kg embedding models \cite[\emph{inter alia}]{bordes2013translating,yang2015embedding,trouillon2016complex}, however, tend to treat each triple independently, which makes multi-hop paths through auxiliary nodes less straightforward to capture. We therefore dereify \ctd during extraction, collapsing each three-triple pattern into a single direct edge labeled with the interaction's action type (e.g., \textit{increases expression}).

\begin{figure*}[t]
    \centering
    \begin{subfigure}[t]{0.5\textwidth}
        \centering
        \includegraphics[width=\linewidth]{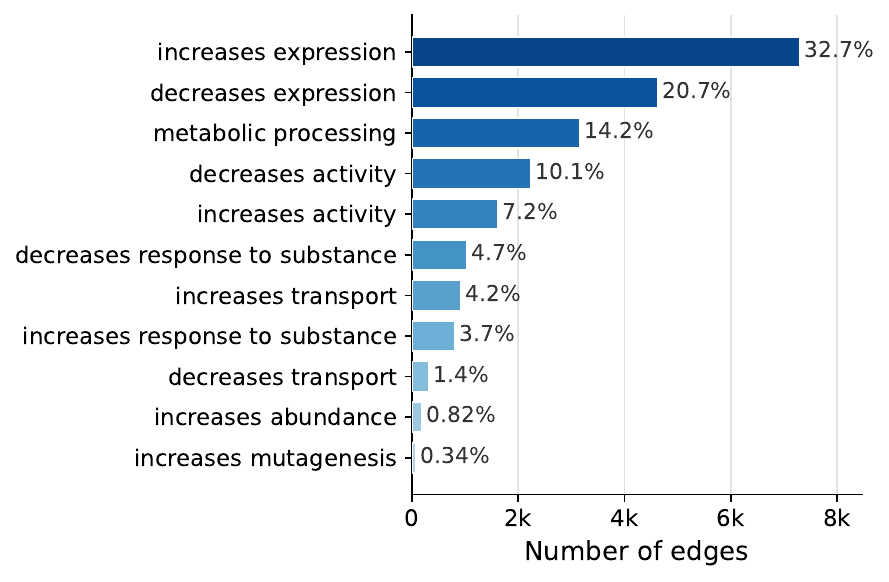}
        \caption{Core \ctd \kg.}
        \label{fig:core_kg_edgetypes}
    \end{subfigure}%
    ~ 
    \begin{subfigure}[t]{0.5\textwidth}
        \centering
        \includegraphics[width=\linewidth]{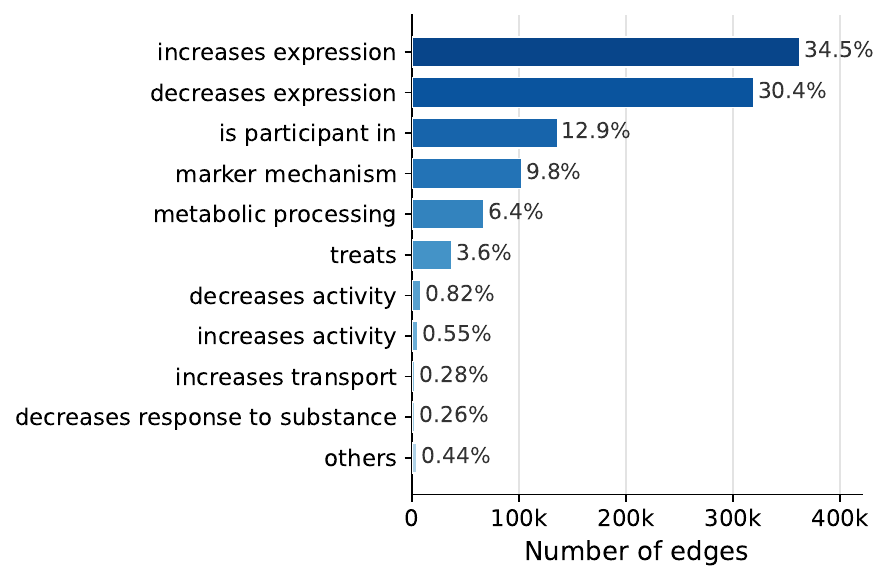}
        \caption{Enriched \ctd \kg.}
        \label{fig:complete_kg_edgetypes}
    \end{subfigure}
    \caption{Distribution of relation types in the (a) \textit{core} and (b) \textit{enriched} \ctd \kgs. "Others" combines 4 relation types from the \textit{enriched} \kg with fewer than 2,600 edges (0.44\%): \textit{increases response to substance}, \textit{decreases transport}, \textit{increases abundance}, \textit{increases mutagenesis}.}
    \label{fig:ctd_kg_rel_dist}
    \vspace{-1em}
\end{figure*}

\rparagraph{Graph construction}
We build the \textit{enriched} \ctd \kg by iteratively growing it outward from the chemical-gene interactions linked to textual evidence (i.e., our \textit{core} set): we attach every chemical-disease, gene-disease, and gene-pathway association whose chemical or gene is already in the graph, so that the added disease and pathway nodes act as hubs linking otherwise-isolated chemicals and genes. We repeat this process until no further connected association can be added or a target graph size is reached.
We finally keep the largest connected component, discarding 10 small isolated ones.
The resulting \textit{enriched} graph has 74,137 
nodes, and 1,048,611 edges spanning 14 relation types. Table \ref{tab:ctd_kg} summarizes its statistics, and Fig. \ref{fig:ctd_kg_rel_dist} shows the distribution of relation types in both \textit{core} and \textit{enriched} \ctd \kg.

\begin{table}[t]
    \centering
    \caption{Statistics of the \textit{core} and \textit{enriched} \ctd \kgs.}
    \label{tab:ctd_kg}

    \resizebox{0.95\textwidth}{!}{%
        \begin{tabular}{lllllll|l}
            \toprule
            & \multirow{2}{*}{\textbf{\# Relations}} & \multirow{2}{*}{\textbf{\# Edges}} & \multicolumn{5}{c}{\textbf{\# Nodes}} \\
            \cmidrule(lr){4-8}
            &  & & Chemicals & Genes & Diseases & Pathways & Total \\
            \midrule
            Core    & 11 & 22,293    & 4,179 & 3,431  & -     & -    & 7,610  \\
            Enriched & 14 & 1,048,611 & 13,773 & 50,709 & 7,292 & 2,363 & 74,137 \\
            \bottomrule
        \end{tabular}%
    }
\end{table}

\subsection{\pubmed Corpus}
\label{sec:pubmed_corpus}

We obtain the documents describing these chemical-gene interactions from \pubmed \cite{lu2011pubmed}, the reference publicly available database of biomedical citations and abstracts maintained by the National Center for Biotechnology Information (NCBI) at the US National Library of Medicine, from which \ctd draws all of its curated interactions.
To obtain the relevant abstracts together with reliable entity annotations, we use PubTator \cite{wei2024pubtator}, an NCBI service that automatically annotates biomedical entities and their relations across \pubmed. We retrieve, using the PubTator 3.0 API, for every PMID recorded in our \ctd interactions, the corresponding abstract and title with its chemical and gene mentions normalized to \ctd identifiers, which allows us to locate the entities of each triple directly in the text.

\subsection{\textit{\ctdalign} Triple-Document Dataset}
\label{sec:ctdalign}
We construct \textit{\ctdalign}, a dataset of one-to-one triple-document pairs used to train and evaluate alignment by pairing \ctd's chemical-gene interactions with their PubMed evidence.\footnote{Data is publicly available at \href{https://github.com/ritatsousa/bio-text-kg-alignment/tree/main/Data/CTD_to_PubTator/kg_data}{https://github.com/ritatsousa/bio-text-kg-alignment/CTD\_to\_PubTator/kg\_data}.}

\rparagraph{Candidate Document Selection}
We start from over 3M raw chemical-gene interactions \ctd records and keep only single-action interactions, discarding the approximately 1.2M multi-action records (40.5\%). A compound action such as \textit{(increases-phosphorylation, decreases-degradation)} has no unambiguous \kg representation, and retaining such records would degrade the semantic quality of the alignment data. 
This filtering results in 1,546,310 unique single-action triples.
Rather than forcing a single paper per triple, which succeeds for only 23.6\% of triples, as papers typically discuss several interactions together, we allow multiple triples per document during collection.
From the 63,274 unique articles referenced by these triples, we download 63,239 abstracts via PubTator and, for each triple, extract the passages mentioning both its chemical and its gene as candidate evidence spans, yielding 45,684 triples with at least one candidate. 

\rparagraph{Document Evidence}
Co-occurrence is necessary but not sufficient, as a sentence may name both entities without describing their relationship (e.g., “Both aspirin and COX-2 were measured at baseline”).
We therefore rank the candidate spans of each triple by semantic relevance. Specifically, we verbalize the triple as a short natural-language query by concatenating its subject name, a human-readable phrasing of the predicate, and its object name (e.g., “Dichlorvos decreases activity of ACHE”), and score every candidate span against this query with a biomedical cross-encoder that applies full cross-attention between query and span.
Concretely, we use MedCPT's re-ranker \cite{jin2023medcpt}, a cross-encoder initialized from PubMedBERT \cite{gu2021domain} and contrastively trained on 18.3M query-article pairs mined from \pubmed search logs.\footnote{In initial experiments comparing MedCPT against BiomedBERT Reranker 2 \cite{neuml2025biomedbert}, a more recent cross-encoder fine-tuned from PubMedBERT, on 1,000 triples sampled with stratified predicate representation, we found that MedCPT retained both entity mentions in 94.8\% of the cases, whereas BiomedBERT dropped at least one entity in 34.4\% of the cases. Hence, we kept MedCPT as our evidence selector.}
For each triple, we select the highest-scoring span as the supporting document.

\rparagraph{One-to-one Mapping}
To obtain a paired dataset, we enforce a strict one-to-one triple-document correspondence: when the same span is selected for several triples, we keep the triple with the least frequent (most informative) predicate and drop the others. 
This produces the \textit{\ctdalign} dataset, sourced from 17,530 source publications and consisting of 22,293 triple-document pairs (794 collisions removed, 3 relation types dropped as a consequence), each comprising a chemical, a gene, a single-action relation, a supporting document (with an average length of 303 characters), and the source PMID for full provenance. Table \ref{tab:ctd_kg} reports the statistics of the corresponding \textit{core} graph.  

\section{Experimental Setup}
\label{sec:exp_setup}

\subsection{Training and Optimization Details}
\label{sec:training_optim_details}

\rparagraph{Knowledge Graph Embeddings}
We experiment with three \kg embedding models that capture different aspects of the graph structure: (i) RotatE~\cite{sun2019rotate}, (ii) TuckER~\cite{balavzevic2019tucker}, and (iii) RDF2Vec~\cite{ristoski2016rdf2vec}.
RotatE \cite{sun2019rotate} embeds entities and relations in a complex vector space and models each relation as an element-wise rotation from the subject to the object entity, thus capturing patterns such as symmetry/asymmetry, inversion, and composition.
TuckER \cite{balavzevic2019tucker} is a bilinear model based on the Tucker decomposition of the binary subject-predicate-object tensor into entity and relation embeddings and a shared core tensor, yielding a fully expressive factorization.
RDF2Vec \cite{ristoski2016rdf2vec}, in contrast, does not optimize a triple-scoring objective, but extracts random walks \cite{de2015substructure} over the graph and trains a Word2Vec model \cite{mikolov2013efficient} on them to obtain entity embeddings.\footnote{In our experiments, we train RDF2Vec only with Skip-gram as it has been shown to yield superior results \cite{ristoski2016rdf2vec}.}

We pretrain all models on the \textit{enriched} \ctd \kg, including the complete set of core triples, to obtain domain-adapted \kg embeddings. 
For all three models we combine entity and relation embeddings into a triple representation using the triple-composition operators from \S\ref{sec:text_kg_embeddings}.\footnote{The single-vector operators (\textit{Hadamard}, \textit{L\textsubscript{1}}, \textit{L\textsubscript{2}}) require equal-length vectors; when entity and relation embeddings differ in size (e.g., TuckER), we truncate them to the smaller common dimensionality.}
We tune the most important hyperparameters of RotatE and TuckER for link prediction using TPE sampling \cite{watanabe2023tree} on 51,289 validation triples (5\% of the \textit{enriched} \ctd \kg), and those of RDF2Vec using grid search on entity-type prediction.
Table \ref{tab:hpo_search_spaces} reports the search spaces and optimal values for each tuned hyperparameter. 
We set the remaining hyperparameters to the values reported in the respective papers, and use the \texttt{gensim} defaults for the Word2Vec Skip-gram model of RDF2Vec.\footnote{\href{https://radimrehurek.com/gensim/models/word2vec.html}{https://radimrehurek.com/gensim/models/word2vec.html}}
We train RotatE and TuckER with Adam \cite{kingma2014adam} for up to 500 epochs, using batch sizes of 512 and 128, and early stopping on validation MRR with a patience of 10 and 5, respectively.

\begin{table}[t]
    \centering
    \caption{Hyperparameter search spaces and best values selected for each model, reported in the format \textit{search space / best value}. We use the following abbreviations: dim. = dimension, self-adv. sampling temp. = the self-adversarial sampling temperature of RotatE \cite{sun2019rotate}. A dash (---) indicates that the hyperparameter was either not tuned (in which case, we report only the value used) or not applicable to the given model.}
    \label{tab:hpo_search_spaces}
    
    \resizebox{\textwidth}{!}{%
    \begin{tabular}{llll}
    \toprule
    \textbf{Hyperparameter} & \textbf{RotatE} & \textbf{TuckER} & \textbf{RDF2Vec} \\
    \midrule
        
        Entity dim. & $\{100, \dots, 500\}$ / 500 & $\{100, 200, 300\}$ / 200 & $\{100, 200, 300\}$ / 200 \\
        Relation dim. & $=$ entity dim & $\{30, 65, 100\}$ / 65 & $=$ entity dim \\
        Learning rate & $[10^{-5}, 5 \times 10^{-4}]$ / $2.8 \times 10^{-4}$ & $[5 \times 10^{-4}, 5 \times 10^{-3}]$ / $5.7 \times 10^{-4}$ & ---  \\
        Margin & $\{3, 6, 9, 12\}$ / 9 & --- & --- \\
        Adv. sampling temp. & $[0.5, 1.0]$ / 0.91  & --- & --- \\
        Negative samples & $[50, 500]$ / 467  & --- & $\{10, 15, 25\}$ / 10  \\
        Dropout (3 layers) & --- & $[0.1, 0.5]$ / 0.22, 0.35, 0.27  & --- \\
        Label smoothing & --- & $\{0.0, 0.1, 0.2\}$ / 0.1  & --- \\
        Walk depth & --- & --- & $\{4, 6, 8\}$ / 4  \\
        Walks / entity & --- & --- & $\{100, 200, 500\}$ / 200  \\
        Word2Vec epochs & --- & --- & $\{10, 25, 50\}$ / 25  \\
    \bottomrule
    \end{tabular}%
    }
\end{table}

\rparagraph{Document Embeddings}
We embed the \pubmed corpus documents (\S\ref{sec:pubmed_corpus}) using two domain-specific pretrained \lms: BioBERT (\texttt{biobert-v1.14}) \cite{lee2020biobert} and PubMedBERT (\texttt{PubMedBERT-base-uncased-abstract}) \cite{gu2021domain}. BioBERT is initialized from BERT \cite{devlin2019bert} and further pre-trained on \pubmed abstracts and \pubmed Central full-text articles, whereas PubMedBERT is trained from scratch exclusively on biomedical corpora. In both cases, we obtain a 768-dimensional document embedding via mean pooling over the token-level representations.\footnote{Note that document embeddings can also be obtained directly from sentence encoders or embedding-capable \llms. Our framework is agnostic to the choice of the text encoder. Here we adopt established, openly available domain-specialized \lms and leave the exploration of other families of document embedders to future work.}

\rparagraph{Alignment Model}
We evaluate three projection heads (cf. \S \ref{sec:alignment}). We run hyperparameter optimization separately for each combination of text encoder, \kg embedding model, projection head, triple-composition operator, hard-negatives sampling strategy, and training direction using TPE sampling. 
We search for the optimal hidden size of the alignment projector in the interval $\{256, 512\}$, finding 256 as the best value. We note that this hyperparameter is not used for the linear projector. 
Similarly, we select a dropout of 0.273 over  the interval $[0.1, 0.3]$, a weight decay of $1.33 \times 10^{-3}$ over the log-scaled interval $[10^{-5}, 10^{-2}]$, and a learning rate  of $1.05 \times 10^{-3}$ over the log-scaled interval $[10^{-4}, 5 \times 10^{-3}]$.
We train the alignment model with the InfoNCE loss \cite{oord2018representation} using in-batch negatives and a temperature of 0.07. Optionally, we add four hard \kg negatives per example, generated with different triple-corruption strategies as per \S \ref{sec:alignment}. 
We optimize using AdamW \cite{loshchilov2019decoupled}, for a maximum of 100 epochs with early stopping (patience of 15 epochs), a batch size of 256, and gradient clipping at 1.0.

\rparagraph{Infrastructure and Compute}
We conduct all training and evaluation on a cluster with virtual machines. We train the RotatE and TuckER \kg embedding models on single NVIDIA A100 (80 GB) and H100 (94 GB) GPUs, and the RDF2Vec model on a machine with 48 CPU cores and 128 GB RAM. We conduct all alignment experiments on CPU, allocating 50 GB of RAM per job.\footnote{Our implementation is publicly available:
\href{https://github.com/ritatsousa/bio-text-kg-alignment}{https://github.com/ritatsousa/bio-text-kg-alignment}.}

\subsection{Evaluation Protocol}
\label{sec:eval}
In addition to the full grid comparison, we include a simple baseline that verbalizes each triple using its subject name, predicate label, and object name (e.g., “Cyclophosphamide metabolic processing CYP2B6”) and embeds the resulting text using the same text encoder used for documents. Candidates are then ranked by cosine similarity to the query: in \texttt{D2T} we rank all verbalized triples against the query document, and in \texttt{T2D} we rank all document embeddings against the verbalized query triple. Thus, this baseline relies solely on the text encoder and uses no \kg embeddings. 
We evaluate using 5-fold cross-validation grouped by \pubmed paper to ensure that documents from the same article never appear in both the training and test sets. Within each fold, we hold out 15\% of the training samples for validation. 
We report the mean and standard deviation across the five folds for standard ranking metrics MRR and Hits$@k$, as well as the median rank of the correct item computed over the full candidate pool. That is, each query is ranked against the entire candidate pool of the target modality, i.e., all \kg triples in \texttt{D2T} and all documents in \texttt{T2D}, rather than against a sampled subset of negatives. 

\section{Results and Discussion}
\label{sec:results_discussion}

We evaluate the quality of alignment as full-pool document-to-triple and triple-to-document retrieval. We first discuss the performance of the best model configurations, comparing the results against the verbalized nearest-neighbor baseline (\S\ref{sec:eval}). Afterwards, we discuss the choice of alignment target space, and finally decompose the remaining dimensions.

\begin{table}[t]
    \centering
    \caption{Retrieval performance in both settings. We report averages and standard deviations over five folds. \textit{Rank} denotes absolute median rank. For each training direction, we report its best-on-average configuration; the baseline uses no \kg embeddings. The \textit{Config} column reports the text encoder and \kg embedding model. All aligned configurations use a linear projection head over concatenated subject, predicate, object embeddings. The best results per column are highlighted in bold, the second best are underlined.}
    \label{tab:main_results}

    \resizebox{\textwidth}{!}{%
        \begin{tabular}{llllllllll}
            \toprule
            
            &  & \multicolumn{4}{c}{\textbf{D2T} (document$\to$triple)} & \multicolumn{4}{c}{\textbf{T2D} (triple$\to$document)} \\
            \cmidrule(lr){3-6}\cmidrule(lr){7-10}
            
            Model & Config & 
            MRR & Hits@1 & Hits@10 & Rank & 
            MRR & Hits@1 & Hits@10 & Rank\\
            \midrule

            Verbalized & BioBERT / --
            & 8.41\textsubscript{$\pm$0.3} & 5.47\textsubscript{$\pm$0.3} & 13.69\textsubscript{$\pm$0.4} & 927\textsubscript{$\pm$45} & \underline{7.41\textsubscript{$\pm$0.1}} & \underline{4.48\textsubscript{$\pm$0.1}} & \underline{12.98\textsubscript{$\pm$0.2}} & 845\textsubscript{$\pm$51} \\

            Verbalized & PubMedBERT / --
            & 4.43\textsubscript{$\pm$0.3} & 2.81\textsubscript{$\pm$0.2} & 7.45\textsubscript{$\pm$0.4} & 4826\textsubscript{$\pm$96} & 2.83\textsubscript{$\pm$0.2} & 1.68\textsubscript{$\pm$0.2} & 4.87\textsubscript{$\pm$0.3} & 5066\textsubscript{$\pm$136} \\

            \hdashline
            
            Aligned (text$\to$\kg) & PubMedBERT / RDF2Vec
            & \textbf{15.99\textsubscript{$\pm$0.6}} & \textbf{8.74\textsubscript{$\pm$0.6}} & \textbf{30.17\textsubscript{$\pm$0.7}} & \textbf{38\textsubscript{$\pm$2}} &
            \textbf{12.13\textsubscript{$\pm$0.3}} & \textbf{5.92\textsubscript{$\pm$0.2}} & \textbf{24.30\textsubscript{$\pm$0.8}} & \textbf{55\textsubscript{$\pm$2}} \\

            Aligned (\kg$\to$text) & BioBERT / RotatE
            & 3.75\textsubscript{$\pm$0.3} & 1.34\textsubscript{$\pm$0.3} & 7.84\textsubscript{$\pm$0.6} & 349\textsubscript{$\pm$18} & 5.89\textsubscript{$\pm$0.2} & 2.21\textsubscript{$\pm$0.3} & 12.41\textsubscript{$\pm$0.3} & \underline{147\textsubscript{$\pm$7}} \\

            Aligned (bidirectional) & PubMedBERT / RotatE
            & \underline{15.15\textsubscript{$\pm$0.6}} & \underline{8.51\textsubscript{$\pm$0.6}} & \underline{28.27\textsubscript{$\pm$0.6}} & \underline{48\textsubscript{$\pm$3}} & 5.74\textsubscript{$\pm$0.3} & 2.29\textsubscript{$\pm$0.3} & 11.97\textsubscript{$\pm$0.5} & 159\textsubscript{$\pm$5} \\    
            
            \bottomrule
        \end{tabular}%
    }
\end{table}

\subsection{Overall Retrieval Performance}
\label{sec:retrieval_performance}

Table \ref{tab:main_results} contrasts the best alignment configuration under each training direction (i.e., under each choice of shared retrieval space) with the verbalization baseline, in both retrieval settings. 
Learning a projection between the two frozen text and \kg embedding spaces substantially outperforms the baseline in both retrieval settings.
The best model projects text into the \kg space with a \textit{linear} head over the \textit{concatenation} of the subject, predicate, and object embeddings; it improves MRR by 90\% on \texttt{D2T} and 64\% on \texttt{T2D} over the stronger (BioBERT) baseline.
The gain is even larger over the full ranking, namely the median rank of the correct items drops from 927 to 38 on \texttt{D2T} (a 24$\times$ improvement), and from 845 to 55 on \texttt{T2D} (a 15$\times$ improvement), while Hits@10 roughly doubles in both settings. 
Performance is also balanced across relation types. For the best model, relation-averaged (macro) MRR exceeds the instance-level (micro) value (18.00 vs. 15.99 on \texttt{D2T}), indicating that the results are not dominated by a few frequent predicates.
Notably, the strongest baseline uses BioBERT, whereas the best alignment model uses PubMedBERT, indicating that once the projection is learned, retrieval becomes largely independent of the text encoder that a purely text-based approach depends on.

\subsection{The Alignment Target Space}
\label{sec:alignment_space}
\begin{figure*}[t]
    \centering
    \begin{subfigure}[t]{0.48\textwidth}
        \centering
        \includegraphics[width=\linewidth]{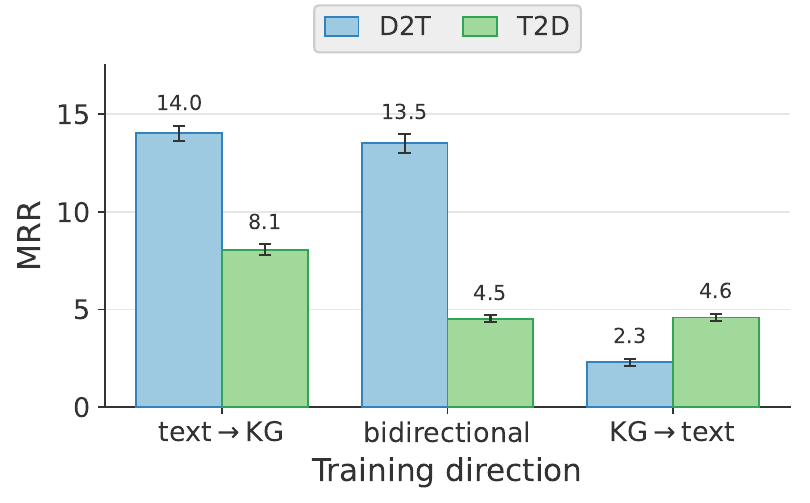}
        \caption{Training direction (\textit{concatenation}, in-batch negatives only).}
        \label{fig:training_direction}
    \end{subfigure}%
    ~ 
    \begin{subfigure}[t]{0.48\textwidth}
        \centering
        \includegraphics[width=\linewidth]{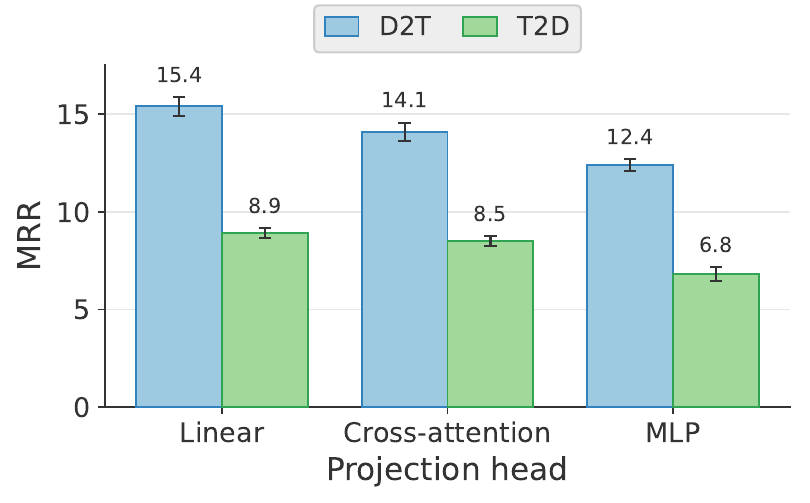}
        \caption{Projection head (within \textit{text$\to$\kg}, \textit{concatenation}).}
        \label{fig:projection_head}
    \end{subfigure}
    ~ 
    \begin{subfigure}[t]{0.48\textwidth}
        \centering
        \includegraphics[width=\linewidth]{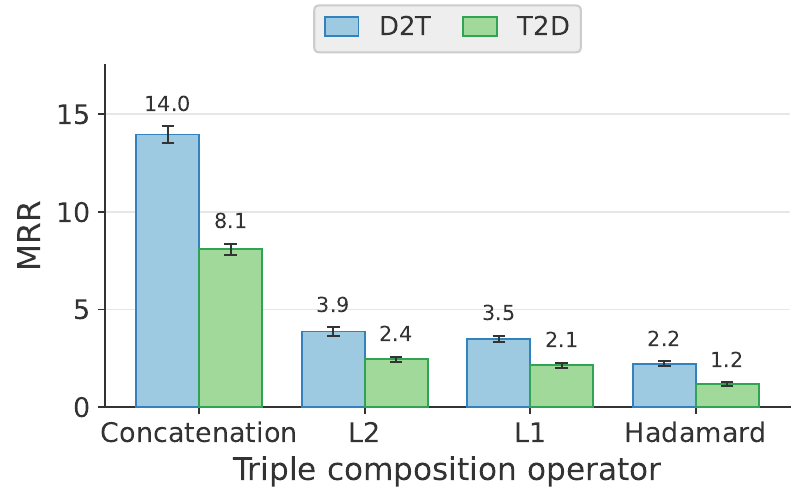}
        \caption{Triple composition overall.}
        \label{fig:triple_composition}
    \end{subfigure}%
    ~ 
    \begin{subfigure}[t]{0.48\textwidth}
        \centering
        \includegraphics[width=\linewidth]{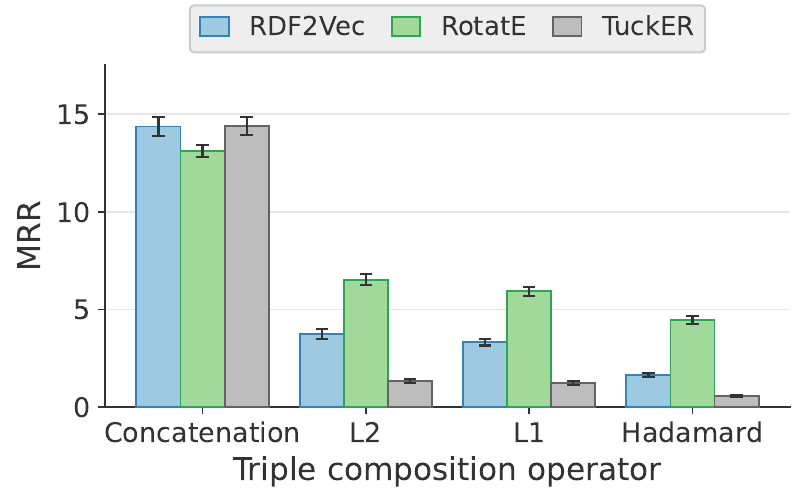}
        \caption{Triple composition by \kg embedding (\texttt{D2T}).}
        \label{fig:kg_composition}
    \end{subfigure}
    \caption{Impact of the training direction, projection head, and triple-composition operator on MRR. Bars show both retrieval settings (\texttt{D2T}, \texttt{T2D}), except (d), which shows \texttt{D2T} only.}
    \label{fig:primary_dimensions}
\end{figure*}

Crucially, projecting text into the \kg space outperforms the other training directions in both evaluation settings. The alternative strategies are strongly asymmetric: the \textit{bidirectional} approach matches text$\to$\kg on \texttt{D2T} (15.15 vs. 15.99 MRR), but collapses on \texttt{T2D} (5.74 vs. 12.13 MRR), while \textit{\kg$\to$text} heavily underperforms in both settings, falling below the strong (BioBERT) verbalization baseline. 
We confirm this ordering through a controlled comparison holding the remaining factors fixed, as shown in Fig. \ref{fig:training_direction}. 

We hypothesize that this asymmetry stems from the two mappings being ill-posed in opposite ways. Projecting text into the \kg space is a \textit{reduction}: a document encodes more information than its triple counterpart, e.g., additional entities, surface form, and syntax. Consequently, the model needs only discard information that becomes redundant once the relation is identified. The reverse mapping is an \textit{addition}: from a compact, decontextualized triple, \textit{\kg$\to$text} must reconstruct precisely this missing context and surface realization. This inverse problem is under-determined, as a single triple may fit many documents, and recovering absent information is inherently harder than discarding it. The geometry of the two spaces reinforces this asymmetry, as pretrained text embeddings are anisotropic, occupying a narrow cone \cite{ethayarajh2019contextual,li2020sentence,gao2021simcse}, so retrieving in text space means separating tightly packed neighbors, whereas the \kg space offers more discriminative targets. Fixing the better-structured space as the target resembles the locked-encoder recipe that has proven effective in cross-modal alignment \cite{zhai2022lit}.
Finally, we find that the \textit{bidirectional} strategy brings no further gains. In each evaluation setting, it mirrors the corresponding single-direction projection (text$\to$\kg for \texttt{D2T} and \textit{\kg$\to$text} for \texttt{T2D}), inheriting the latter's weakness. Consequently, \texttt{T2D} is still ranked in the less discriminative text space.

\subsection{Impact of Design Dimensions}
\label{sec:dimension_impact}
We next rank the six design dimensions by their impact on performance. For each factor, we average MRR over all configurations that share a given level of that factor (its marginal effect) and quantify the factor's share of the variance with a one-way $\eta^2$ (i.e., the between-group over total sum of squares across configurations) on the balanced grid combining the \textit{text$\to$\kg} and \textit{bidirectional} training directions.
Almost all of the variance can be attributed to a single factor. Specifically, the triple composition alone explains 87\% of the \texttt{D2T} and 62\% of the \texttt{T2D} MRR variance, and the \kg embedding a further 6\% and 3\%, respectively. 
The projection head, text encoder, and negative sampling scheme, have a negligible impact, each explaining less than 1\%.
Note that per-factor main effects do not need to sum to one. The remainder reflects interactions, particularly between \kg embedding and triple composition (see below), and residual variance. 
Therefore, we first analyze the effect of the triple-composition operator, then fix it to its best value (\textit{concatenation}) and study the remaining factors as \textit{conditional} effects.

\rparagraph{Triple Composition}
\textit{Concatenation} substantially outperforms the other operators (Fig. \ref{fig:triple_composition}): within \textit{text$\to$\kg} it reaches 14.0 MRR on \texttt{D2T}, compared to 3.9 or less for \textit{L\textsubscript{1}}, \textit{L\textsubscript{2}}, and \textit{Hadamard} operators.
Concatenation preserves the separable subject-predicate-object structure in a $3d$ vector, whereas the other three operators collapse the triple into a $d$-dimensional vector -- the \textit{Hadamard product} via a fixed multiplicative form (as in DistMult \cite{yang2015embedding}), and the \textit{L\textsubscript{1}/L\textsubscript{2}} operators via a translation one (as in TransE \cite{bordes2013translating}) -- potentially discarding the information needed by the projection to recover the relation.\footnote{We note that concatenation's advantage is inseperable from its larger output dimensionality.}
This gap can also be observed in the baseline comparison, where only the concatenation-based models exceed the verbalization baseline. Alignment quality thus depends heavily on how the triple representation is composed.

\rparagraph{Interaction of \kg Embedding and Triple Composition}
The quality of a \kg embedding is tightly intertwined with the triple-composition operator (Fig. \ref{fig:kg_composition}). 
Averaged over all operators, RotatE appears best, but only because it is the least harmed by the poorly performing operators (i.e., averaging artifact). 
Under \textit{concatenation}, the ranking reverses, namely RDF2Vec and TuckER (both 14.4 MRR) outperform RotatE (13.1 MRR) for \texttt{D2T}, and the gap widens sharply for \texttt{T2D}, where RotatE collapses to 4.2 MRR against RDF2Vec's 10.6 MRR.  Marginal averages over the composition operator are thus misleading. 
RDF2Vec, built from Word2Vec over graph walks \cite{ristoski2016rdf2vec}, yields a distributional geometry closest to that of \lms (i.e., semantically meaningful relation embeddings). This plausibly makes it the easiest and most robust alignment target across retrieval settings. In contrast, TuckER's tensor-factorization geometry \cite{balavzevic2019tucker} and RotatE's rotational one \cite{sun2019rotate} are optimized for link prediction rather than cosine similarity with text.

\begin{figure*}[t]
    \centering
    \begin{subfigure}[t]{0.48\textwidth}
        \centering
        \includegraphics[width=\linewidth]{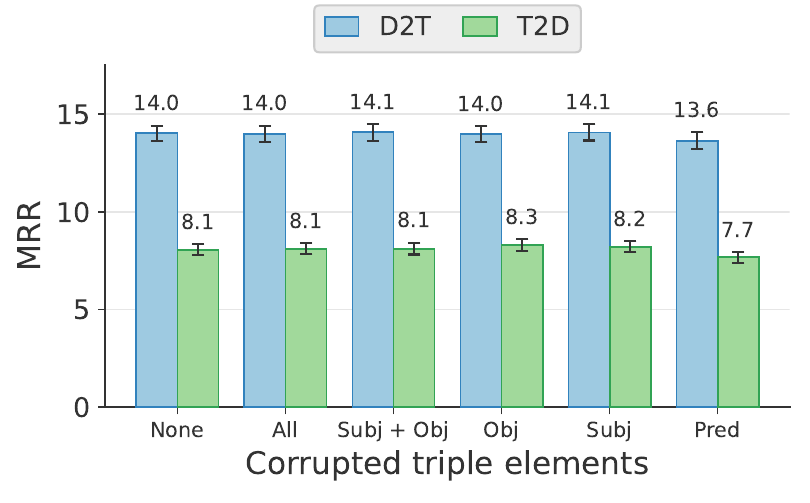}
        \caption{Hard-negatives selection (within \textit{text$\to$\kg}, \textit{concatenation}).}
        \label{fig:hard_negatives}
    \end{subfigure}
     ~
    \begin{subfigure}[t]{0.48\textwidth}
        \centering
        \includegraphics[width=\linewidth]{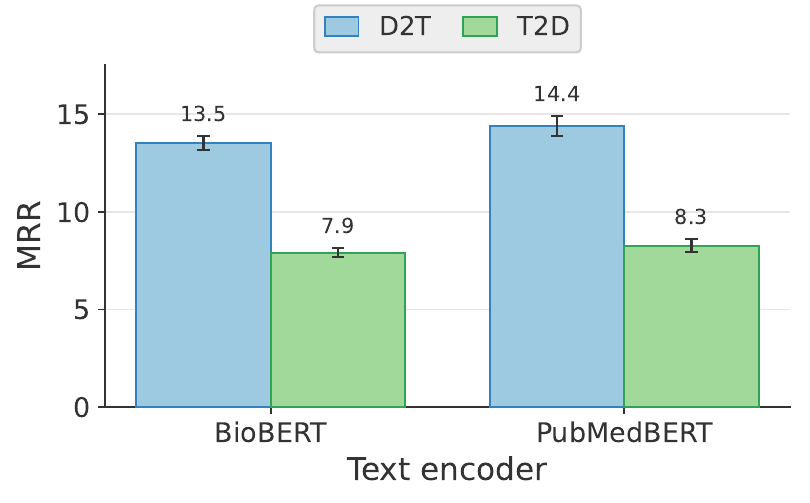}
        \caption{Text encoder (within \textit{text$\to$\kg}, \textit{concatenation}).}
        \label{fig:text_encoder}
    \end{subfigure}%
    \caption{Impact of hard-negatives selection and the text encoder on MRR, in both retrieval settings.}
    \label{fig:secondary_dimensions}
    \vspace{-1em}
\end{figure*}

\subsection{Secondary Design Choices}
\label{sec:secondary_choices}

Fixing triple composition to \textit{concatenation}, we find that the remaining within-setup choices have small and often negligible effects.

\rparagraph{Projection Head}
The \textit{linear projection} outperforms both the \textit{MLP} and the \textit{cross-attention} in both retrieval settings (Fig. \ref{fig:projection_head}). 
The marginal view hides this effect entirely, as the projection head explains under 0.4\% of MRR variance when averaged over all operators. 
However, once the triple composition is fixed to \textit{concatenation}, the projection head becomes the single largest factor on \texttt{D2T}, explaining 69\% of the variance among concatenation-based configurations. On \texttt{T2D}, it is second only to the \kg embedding, whose choice dominates because of the RotatE collapse discussed above.

\rparagraph{Hard Negative Sampling Scheme}
Fig. \ref{fig:hard_negatives} shows that contrasting the paired document and triple against in-batch negatives suffices for training the alignment model. No hard negative selection strategy based on corrupting the triple provides any meaningful benefits over using none, in 
either \textit{text$\to$\kg} or \textit{bidirectional} training directions.\footnote{By construction, \textit{KG$\to$text} uses in-batch negatives only and is excluded from this comparison.}
The only visible deviation is \textit{predicate-only corruption}, which performs slightly worse in both settings.

\rparagraph{Text Encoder}
Lastly, the choice between BioBERT and PubMedBERT has only a small impact once the projection is learned (Fig. \ref{fig:text_encoder}). This contrasts sharply with the verbalization baseline,
where BioBERT far outperforms PubMedBERT (e.g., 8.41 vs. 4.43 MRR on \texttt{D2T}, 7.41 vs. 2.83 on \texttt{T2D}). After alignment, this gap reverses, as PubMedBERT performs slightly better and yields the best overall configuration. This points to the fact that most of the encoder-level difference observed in the baseline could be explained by the verbalized representation rather than the encoder itself, and is rendered largely inconsequential once a projection is learned. 

\section{Limitations and Future Work}
\label{sec:limitations}

Our analysis trades breadth for a controlled comparison of design dimensions. 
First, we evaluate a single relation family (chemical-gene interactions) from one resource (\ctd). Replicating the comparison on broader biomedical \kgs with richer relation sets (e.g., Hetionet \cite{himmelstein2017systematic}, PrimeKG \cite{chandak2023building}), would test whether the findings transfer to other relation types.
Second, we restrict evaluation to one-to-one triple-document pairs, whereas in practice a document may describe several interactions (one-to-many) and multiple documents may support the same triple (many-to-one). This isolates the core alignment signal and simplifies evaluation, but it also coincides with the single-positive assumption of InfoNCE \cite{oord2018representation}. A natural extension is many-to-many alignment via a multi-positive (supervised) contrastive objective \cite{khosla2020supervised} that treats all valid document-triple matches as positives. 
Finally, we assess alignment intrinsically through retrieval. Evaluating the learned representations on a downstream task, such as \kg completion or evidence-grounded retrieval, would more directly measure their practical utility.

\section{Conclusion}
\label{sec:conclusion}

We introduce a unified framework to systematically compare lightweight strategies for aligning biomedical text and \kgs across six key design dimensions: text encoder, \kg embedding model, triple-composition operator, projection head, training direction, and hard-negatives selection strategy. 
Through extensive retrieval experiments on \textit{CTD-Align}, a new corpus of over 22K one-to-one chemical-gene triple-document pairs, we find that the triple-composition operator and the shared retrieval space (i.e., training direction) have the greatest impact, whereas the text encoder and hard-negatives selection matter comparatively little for retrieval performance.
Simpler approaches consistently outperform more complex ones: projecting text into the \kg space with a linear head over the concatenated subject, predicate, and object embeddings yields the best alignment.
We hope these findings will inspire more transparent evaluation, including systematic ablations that uncover the components driving performance.

\begin{acknowledgments}
The work presented in this paper has been partly funded by the German Federal Ministry of
Education and Research under grant number 13GW0661C (KI-DiabetesDetektion).
The authors also acknowledge support by the state of Baden-Württemberg through bwHPC.
\end{acknowledgments}

\section*{Declaration on Generative AI}
During the preparation of this work, the authors used Opus 4.8 and Codex (GPT 5.5) in order to: generate images, generate the abstract, paraphrase and reword, improve the writing style, grammar and spelling check, peer review simulation, and code support. After using these tools, the authors reviewed and edited the content as needed and take full responsibility for the publication's content. 

\bibliography{references}

\end{document}